\documentclass[letterpaper, 10 pt, conference]{ieeeconf}  

\IEEEoverridecommandlockouts                              

\usepackage{amsmath, theorem}
\usepackage{amssymb}

\usepackage[geometry]{ifsym}
\usepackage{cite}
\usepackage{graphicx}
\newsavebox\mybox

\usepackage{ifsym}

\usepackage{caption}
\usepackage{subcaption}
\usepackage{array}
\usepackage[hidelinks]{hyperref}

\usepackage{enumerate}
\let\labelindent\relax
\usepackage[inline]{enumitem}
\usepackage[ruled,linesnumbered]{algorithm2e}
\usepackage{ stmaryrd }
\usepackage{mathtools}

\usepackage[acronym]{glossaries}
\newacronym{gr1}{GR(1)}{Generalized Reactivity(1)}
\newacronym{ltl}{LTL}{Linear Temporal Logic}
\newacronym{sltl}{sLTL}{Simple Linear Temporal Logic}
\newacronym{ec}{EC}{Environment Characterization}
\glsdisablehyper

\newenvironment{flushitemize}{%
\begin{list}{$\bullet$}
  {\setlength{\leftmargin}{15pt}}%
    \setlength{\labelwidth}{20pt}
    \setlength{\itemindent}{0pt}
    \setlength{\labelsep}{0.5em}
 \setlength{\itemsep}{1pt}
 \setlength{\parskip}{0pt}
 \setlength{\parsep}{0pt}}
 {\end{list}}

\usepackage[normalem]{ulem}
\usepackage{color}
\usepackage{xcolor}

\newcommand{\call}{\mathcal}

\def\cA{{\call A}}

\def\cI{{\call I}}

\def\cM{{\call M}}

\def\cO{{\call O}}

\def\cR{{\call R}}

\def\cX{{\call X}}
\def\cY{{\call Y}}

\newcommand{\envspec}{\varphi_\textrm{e}}
\newcommand{\sysspec}{\varphi_\textrm{s}}
\newcommand{\envinit}{\varphi_\textrm{e}^\textrm{i}}
\newcommand{\sysinit}{\varphi_\textrm{s}^\textrm{i}}
\newcommand{\envsafety}{\varphi_\textrm{e}^\textrm{t}}
\newcommand{\envsafetyvio}{\varphi_\textrm{violated}}
\newcommand{\envsafetynothard}{\varphi_\textrm{e}^\textrm{t,skill}}
\newcommand{\envsafetyhard}{\varphi_\textrm{e}^\textrm{t,hard}}

\newcommand{\syssafety}{\varphi_\textrm{s}^\textrm{t}}

\newcommand{\envlive}{\varphi_\textrm{e}^\textrm{g}}
\newcommand{\syslive}{\varphi_\textrm{s}^\textrm{g}}

\newcommand{\spec}{\varphi}
\newcommand{\spectask}{\spec_{\textrm{task}}}

\newcommand{\psicheck}{\psi_i^{\textrm{check}}}
\newcommand{\envrelaxhard}{\psi_{\textrm{r}}^{\textrm{hard}}}
\newcommand{\envrelaxnothard}{\psi_{\textrm{r}}^{\textrm{skill}}}

\newcommand{\inp}{\cX}
\newcommand{\inpsyms}{\inp_\textrm{c}}
\newcommand{\inpuser}{\inp_\textrm{u}}
\newcommand{\inpuserprov}{\inp_\textrm{user}}
\newcommand{\out}{\cY}
\newcommand{\outnew}{\out_{\textrm{new}}}

\newcommand{\statevar}{\sigma}
\newcommand{\statevarcontrol}{\statevar_\textrm{c}}
\newcommand{\var}{\pi}
\newcommand{\varol}[2]{\pi_{{#1}}^{{#2}}}

\newcommand{\xinitrobot}{x_{\textrm{init}}^{\textrm{robot}}}
\newcommand{\xinitobs}{x_{\textrm{init}}^{\textrm{obs}}}

\newcommand{\inpstate}{\statevar_{\inp}}
\newcommand{\inpstateinit}{\statevar_{\textrm{init}}}
\newcommand{\outstate}{\statevar_{\out}}

\newcommand{\inpstateprime}{\statevar_{\inp}'}

\newcommand{\inpstateone}{\statevar_{\inp, \textrm{1}}}

\newcommand{\inpstatetwo}{\statevar_{\inp, \textrm{2}}}

\newcommand{\inpstatethree}{\statevar_{\inp, \textrm{3}}}

\newcommand{\vars}{\mathcal{V}}
\newcommand{\setor}{~|~}

\newcommand{\true}{{\tt{True}}}
\newcommand{\false}{{\tt{False}}}

\newcommand{\strategy}{\cA}
\newcommand{\monitor}{\cM}

\newcommand{\grounding}{\textrm{G}}
\newcommand{\inversegrounding}{\grounding^{-1}}
\newcommand{\norm}[1]{\left\lVert#1\right\rVert}

\newcounter{examplecounter}
\newcommand{\examplelabel}[1]{%
  \refstepcounter{examplecounter}%
  \label{#1}%
  \theexamplecounter%
}

\newcommand{\exampletwolabel}[1]{%
  \refstepcounter{examplecounter}%
  \label{#1}%
  \theexamplecounter%
}

\newcounter{problemcounter}
\newcommand{\problemlabel}[1]{%
  \refstepcounter{problemcounter}%
  \label{#1}%
  \theproblemcounter%
}

\title{\LARGE \bf
Automated Robot Recovery from Assumption Violations\\
of High-Level Specifications
}

\author{Qian Meng and Hadas Kress-Gazit
\thanks{The authors are with Cornell University in Ithaca, NY 14850, USA. Email: {\tt\small \{qm34,hadaskg\}@cornell.edu}. 
This work is supported by ONR PERISCOPE MURI award N00014-17-1-2699.
}}

\begin{document}
\maketitle

\begin{abstract}
This paper presents a framework that enables robots to automatically recover from assumption violations of high-level specifications during task execution. 
In contrast to previous methods relying on user intervention to impose additional assumptions for failure recovery, 
our approach leverages synthesis-based repair to suggest new robot skills that, when implemented, repair the task.  
Our approach   
detects violations of environment safety assumptions during the task execution, 
relaxes the assumptions to admit observed environment behaviors, and 
acquires new robot skills for task completion. 
We demonstrate our approach with a Hello Robot Stretch in a factory-like scenario. 
\end{abstract}
\section{Introduction}
Formal synthesis -- the process of transforming a high-level specification into a correct-by-construction system -- is increasingly being used to create provably-correct robot behaviors that fulfill complex, high-level tasks~\cite{kress2018synthesis}. 
A common approach to synthesis first creates a discrete abstraction of the environment and the robot skills (or actions)~\cite{he2018automated, silver2023predicate, konidaris2018skills, ahmetoglu2022deepsym}, 
then generates a temporal logic~\cite{clarke2000model} specification from the abstraction and the user-provided tasks. 
Synthesis algorithms (e.g.~\cite{bloem2012synthesis}) transform the specification into a symbolic strategy that guides the composition of the robot skills to satisfy the high-level task. 
If the synthesis fails, meaning the task cannot be guaranteed, researchers have leveraged the feedback from synthesis to modify the abstraction and the specification to make the specification realizable~\cite{fainekos2011revising, li2011mining, raman2012explaining, guo2013revising, alur2013counter}. 

Synthesis for \gls{gr1} specifications has been used for synthesizing robot control from high-level specifications due to its relatively-low quadratic time complexity in the size of its state space~\cite{bloem2012synthesis}. 
\gls{gr1} specifications are a fragment of \gls{ltl}~\cite{clarke2000model} and are structured as an implication from 
\textit{assumptions} on the environment behaviors to \textit{guarantees} on the system behaviors, as described in 
Section~\ref{sec:synth}.
The assumptions and guarantees contain \textit{safety constraints}, which specify requirements that should always hold, and \textit{liveness constraints}, which specify requirements that should hold repeatedly.  
Because of the implication, the synthesized strategy is only required to satisfy the system guarantees if the environment behaviors satisfy the assumptions.

While \gls{gr1} synthesis can produce provably-correct robot strategies, the high-level specification includes assumptions about the behavior of the environment,
such as the location of obstacles or the state of objects; 
if, at runtime, these assumptions are violated, the behavior of the robot is no longer guaranteed to satisfy the task. 
In this work we address the problem of \textit{autonomously repairing such violations}, if possible, to ensure the robot satisfies its intended task.

\begin{figure}
    \centering
    \includegraphics[width=0.45\textwidth]{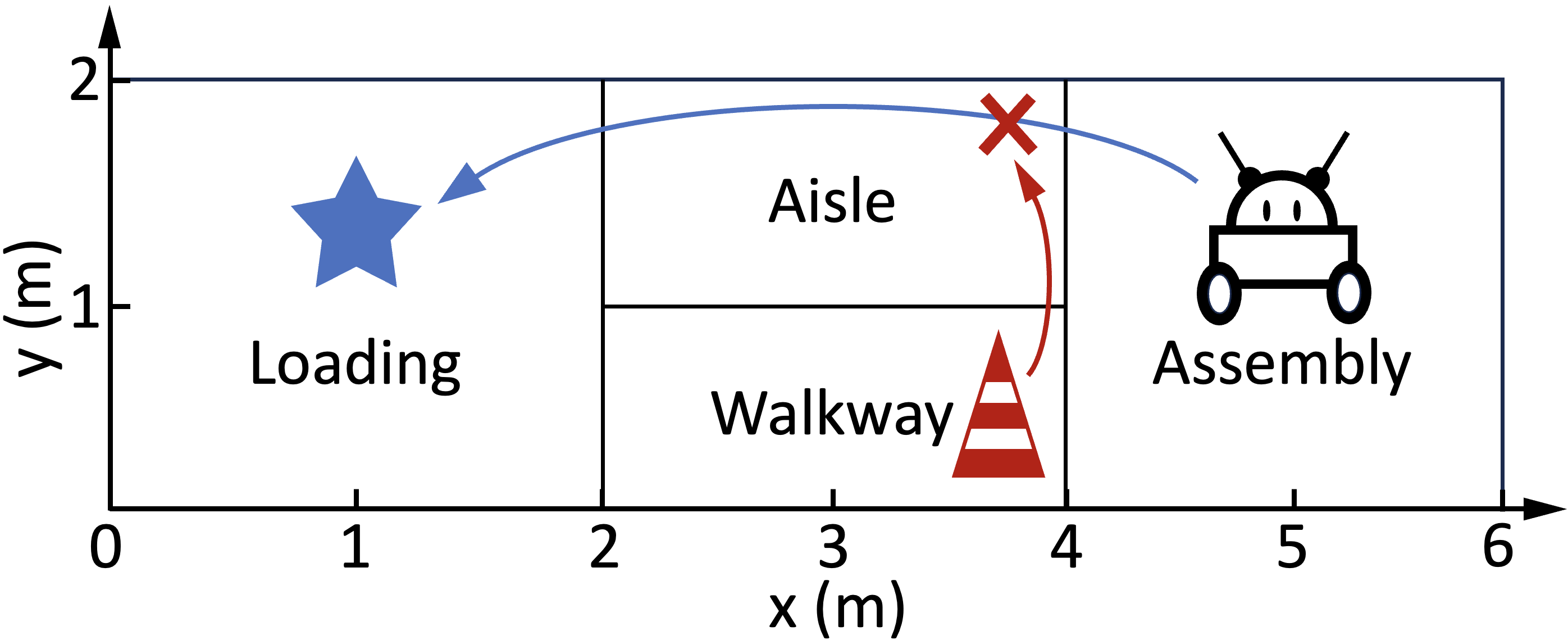}
    \caption{Workspace of Example~\ref{example:1}. 
    }
    \label{fig:example_one_workspace}
\end{figure}

\noindent\textbf{\textit{Example 
\examplelabel{example:1}.}} 
Consider the workspace in Fig.~\ref{fig:example_one_workspace} with four regions. A robot, initially in \emph{Assembly}, is asked to move to \emph{Loading} while ensuring that it is not in the same region as an obstacle, initially located in \emph{Walkway}. 
The robot has a skill $move_1$ which moves the robot from \emph{Assembly} to \emph{Loading} through \emph{Aisle}. 
The user provides the assumption that the obstacle is static and the synthesized strategy contains the action $move_1$. 
During the skill execution when the robot is still in \emph{Assembly}, the obstacle moves to \emph{Aisle}, violating the static-obstacle assumption -- this may result in the robot violating the requirement that it is not in the same region as the obstacle. To repair, we would want the robot to have a skill that goes through \emph{Walkway} instead of through \emph{Aisle}.

\noindent \textit{\textbf{Related work.}} 
Cooperative synthesis creates systems that do not attempt to violate the environment assumptions~\cite{ehlers2015synthesizing,majumdar2019environmentally}. 
Jones et al. 
propose to detect system anomalies by learning Signal Temporal Logic specifications from data and using them to monitor system behaviors~\cite{jones2014anomaly, kong2016temporal}. 
Under topological changes such as the discovery of unknown obstacles, some approaches
modify the abstraction and explore local resynthesis~\cite{livingston2012backtracking, livingston2013patching} and replanning~\cite{lahijanian2016iterative, kantaros2020reactive} for recovery.  
This paper unifies and extends the scope of 
\cite{ehlers2015synthesizing, majumdar2019environmentally, jones2014anomaly, kong2016temporal, 
livingston2012backtracking, livingston2013patching,
lahijanian2016iterative, kantaros2020reactive}.
We attempt to recover from general assumption violations that can arise from
\begin{enumerate*}[label=(\roman*),ref=\roman*]
    \item non-cooperative systems, 
    \item system anomalies such as robot skill failures,
    \item topological changes, and
    \item uncontrollable environment behaviors, such as unexpected changes in user inputs, which were not addressed by 
\cite{ehlers2015synthesizing, majumdar2019environmentally, jones2014anomaly, kong2016temporal, 
livingston2012backtracking, livingston2013patching,
lahijanian2016iterative, kantaros2020reactive}.
\end{enumerate*}

Wong et al.
present a semi-automatic recovery method to tackle general assumption violations for mobile robots~\cite{wong2014correct, wong2018resilient}.
The method a-priori synthesizes a strategy that tolerates \textit{temporary} assumption violations, if possible.
At runtime, the method relaxes the environment assumptions to include new environment behavior but requires user input if the new specification is not realizable. 
Baran et al. propose a human-in-the-loop system that
leverages inverse reinforcement learning to adapt plans according to user preferences~\cite{baran2021ros}.

\noindent\textbf{\textit{Example~\ref{example:1} continued.}} 
In contrast to our approach, the approach in~\cite{wong2014correct, wong2018resilient} recovers from the assumption violation in Example~\ref{example:1} by requiring the user to provide a liveness assumption which states that repeatedly the obstacle is not in \emph{Aisle}.
The robot then waits in \emph{Loading} until the obstacle moves away and then resumes the execution of $move_1$.

This paper aims to recover from assumption violations during runtime automatically. 
We consider a setting where obtaining all feasible robot skills is intractable.  
Therefore, we assume the user provides a set of skills sufficient to satisfy the task under the original environment assumptions.
The process of assumption relaxation 
allows the environment to have more behaviors, but those may prevent the robot from achieving its task.
Rather than weaken the environment by introducing additional environment assumptions as in~\cite{wong2014correct, wong2018resilient}, we extend the robot's capabilities by automatically suggesting and adding new skills that enable it to fulfill its task under the new environment assumptions. 
Recently, \cite{pacheck2020finding, pacheck2022automatic, pacheck2022physically} present a technique for offline repair of an unrealizable specification by suggesting robot skills guaranteed to make the specification realizable. 
We utilize repair to discover new robot skills to complete the task after assumption violations. 

\begin{figure}[t] 
  \centering 
  \includegraphics[width=0.48\textwidth]{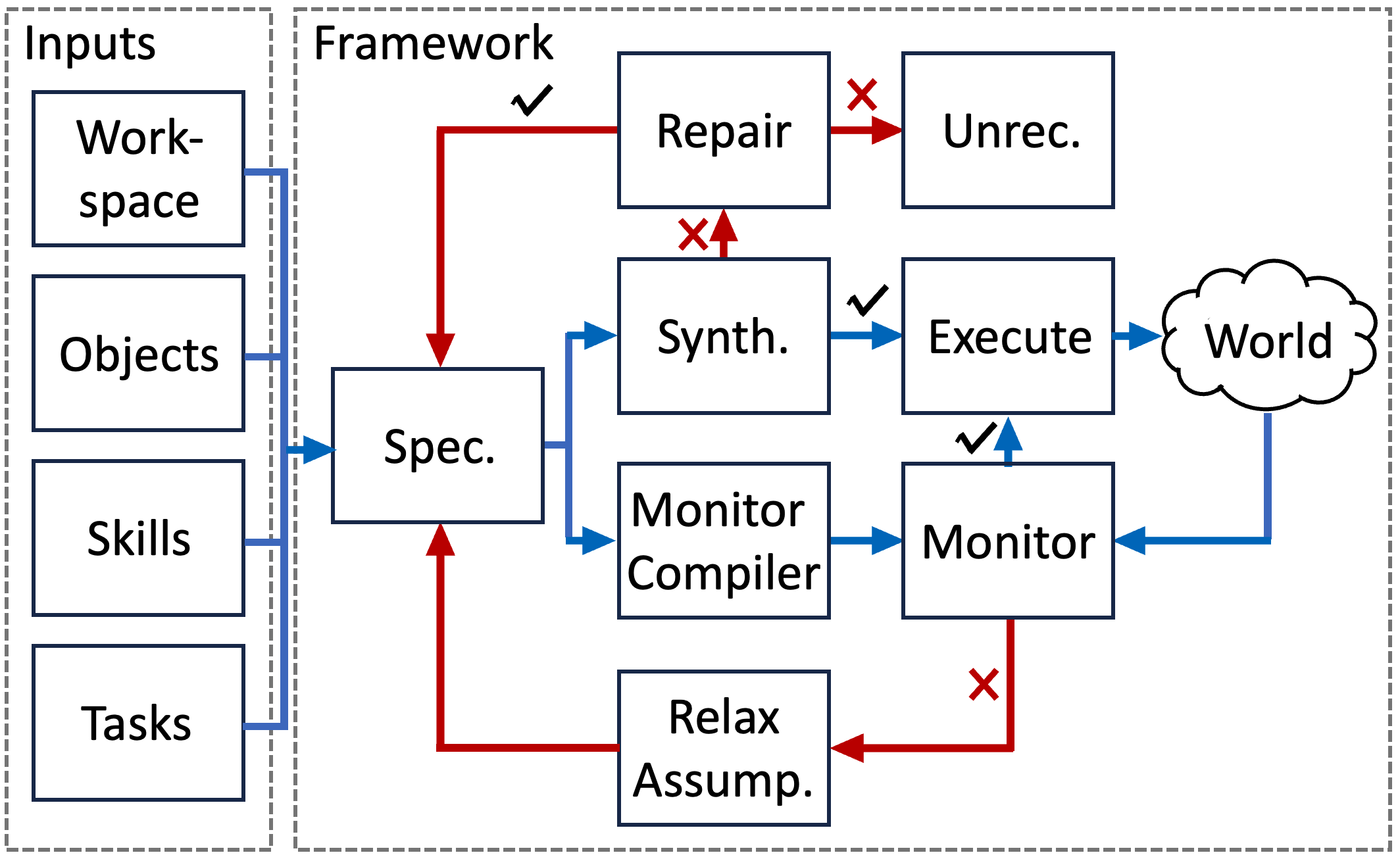} 
  \caption{Our framework for 
  automatic recovery from assumption violations during task execution.
  Blue arrows represent regular execution; 
  red arrows represent our recovery process.
  }
  \label{fig:workflow} 
\end{figure}

This paper presents a framework,
depicted in Fig.~\ref{fig:workflow},
that detects environment assumption violations and leverages the repair tool for recovery. 
We create a specification from the user-provided tasks and an abstraction consisting of the workspace, the objects in the workspace, and the robot skills. 
We automatically create an online monitor for the environment safety assumptions 
(Section~\ref{sec:detection}). 
In response to assumption violations, we relax the violated assumptions (Section~\ref{sec:relaxation}) and perform  
repair to add a set of new skills that enable the robot to complete the task (Section~\ref{sec:repair}). 

\noindent\textbf{\textit{Example~\ref{example:1} continued.}} 
Our approach detects the assumption violation in Example~\ref{example:1} and relaxes the static-obstacle assumption to include the possibility of the obstacle moving from \emph{Walkway} to \emph{Aisle}. 
The specification becomes unrealizable after the assumption relaxation. 
We automatically create a new skill $move_2$ which moves the robot from \emph{Assembly} to \emph{Loading} through \emph{Walkway}. 
We synthesize a new strategy that executes $move_2$ to bring the robot to \emph{Loading}.

\textit{\textbf{Contributions.}}  
To our knowledge, this is the first framework to allow a robot to autonomously recover from assumption violations of temporal logic specifications during task execution. 
This work closes the loop of executing correct-by-construction robot strategies in uncertain environments, for which the initial environment assumptions may be inaccurate, by gradually adjusting the assumptions to meet the actual environment behaviors
and acquiring the required robot skills to satisfy the tasks. 
This work will enable formal synthesis to create autonomous robotic systems that are more applicable to complex, real-world environments.
\section{Preliminaries}

\subsection{Linear Temporal Logic}
We use a fragment of \gls{ltl} to capture specifications. 
Let $\vars$ be a set of atomic propositions. 
We define the syntax of \gls{ltl} recursively using the following grammar:
\begin{equation*}
    \spec \coloneqq  \pi \setor \neg\spec \setor \spec \vee \spec \setor \bigcirc \spec \setor \spec \ \mathcal{U} \  \spec
\end{equation*}
where $\pi\in\vars$ is an atomic proposition, 
$\spec$ is an \gls{ltl} formula, 
$\neg$ is negation, 
$\vee$ is disjunction, 
$\bigcirc$ is the next operator, 
and $\mathcal{U}$ is the strong until operator. 
Conjunction ($\land$), implication ($\to$), equivalence ($\leftrightarrow$), eventually ($\lozenge$), and always ($\square$) can be derived from the grammar. 
 



The semantics of \gls{ltl} are defined over infinite sequences $\sigma = \statevar_0 \statevar_1 \dots$ where $\statevar_i \subseteq \vars$ is the set of atomic propositions that are $True$ at step $i$. 
We denote that a sequence $\sigma$ satisfies a formula $\spec$ at step $i$ by $\sigma^i \models \spec$. 
A sequence $\sigma$ satisfies a formula $\spec$, denoted $\sigma \models \spec$, iff $\sigma^0 \models \spec$. 
Formally, 
$\sigma^i \models \pi $ iff $\pi \in \statevar_i$, 
$\sigma^i \models \bigcirc\spec$ iff $\sigma^{i+1} \models \spec$, 
$\sigma^i \models \lozenge\spec$ iff $\exists j \ge i$ such that $\sigma^j \models \spec$, and 
$\sigma^i \models \square\spec$ iff $\forall j \ge i$, $\sigma^j \models \spec$. 
The full semantics of \gls{ltl} can be found in~\cite{clarke2000model}.

\subsection{Abstractions}
We partition the set of atomic propositions into sets of inputs $\inp$ and outputs $\out$ ($\vars = \inp \cup \out$). 
The inputs describe the environment states, such as the position of an object. 
An input state is a subset of the inputs ($\inpstate \subseteq \inp$). 
We partition the set of inputs into sets of controllable inputs $\inpsyms$ and uncontrollable inputs $\inpuser$ ($\inp = \inpsyms \cup \inpuser$).
Controllable inputs describe the part of the physical world that the robot can control, for example its position. 
A controllable input state is a subset of the controllable inputs ($\statevarcontrol \subseteq \inpsyms$).
Uncontrollable inputs describe the part of the physical world that the robot does not have control over, such as the position of an obstacle, or a user input. 
The input abstraction can be obtained automatically~\cite{he2018automated,silver2023predicate}. 
In this paper, we obtain $\inp$ by taking a Cartesian product of a set of regions $\cR$ that partitions the workspace of the robot and a set of objects $\cO$ that includes the robot,
along with a set of user-controlled inputs $\inpuserprov \subseteq \inpuser$. The set of inputs is
$
    \inp \coloneqq \{\varol{o}{r} \mid o\in\cO, r\in\cR\} \cup \inpuserprov
$ where $\varol{o}{r}$ is $True$ when object $o$ is in region $r$. 
In Example~\ref{example:1}, 
$\cO = \{\textrm{robot}, \textrm{obs}\}$,
 $\cR = \{\textrm{assm}, \textrm{aisle}, \textrm{wlkwy}, \textrm{load}\}$, 
 $\inpuserprov = \emptyset$, 
 $\inpsyms = \{\varol{robot}{r} \mid r \in \cR\}$, 
 $\inpuser = \{\varol{obs}{r} \mid r \in \cR\} \cup \inpuserprov$, 
and $\inp = \inpsyms \cup \inpuser$. 

A grounding function $\grounding$ maps an input proposition to a set of states in the physical world $x = (x_1, \dots, x_n) \in X \subseteq \mathbb{R}^n$, i.e. $\grounding: \inp \to 2^{X}$. 
We lift the grounding function to map an input state $\inpstate$ to a set of physical states by taking the intersection of the groundings of each input in $\inpstate$, 
i.e.  $\grounding(\inpstate) = \bigcap_{\var \in \inpstate} \grounding(\var) \bigcap_{\var \in \inp \setminus \inpstate} (X \setminus \grounding(\var))$. 
An inverse grounding function $\inversegrounding: X \to 2^\inp$ maps a physical state $x \in X$ to an input state $\inpstate \subseteq \inp$ such that 
$x \in \grounding(\inpstate)$
and $\forall \pi\not\in\inpstate, x \not\in \grounding(\pi)$.
In Example~\ref{example:1}, if initially the robot is at $\xinitrobot = (5, 1)$, and the obstacle is at $\xinitobs = (3.8, 0.5)$, then the initial abstract input state is   $ \inpstateinit=\inversegrounding((\xinitrobot,\xinitobs)) = \{\varol{robot}{assm}, \varol{obs}{wlkwy}\}$.  

The set of outputs $\out$ represents the set of skills the robot can execute. 
A skill can only be executed at a set of controllable input states (initial preconditions), 
results in a set of controllable input states (final postconditions), 
and may pass through a set of unique controllable input states (intermediate states). 
Each state except for the final postconditions has a set of postconditions. 
Each state except for the initial preconditions has a set of preconditions.
In Example~\ref{example:1}, the initial precondition of the skill $move_1$ is 
$\inpstateone = \{\varol{robot}{assm}\}$, 
the postcondition of $move_1$ on $\inpstateone$ is 
$\inpstatetwo = \{\varol{robot}{aisle}\}$, and
the postcondition of $move_1$ on $\inpstatetwo$ is 
$\inpstatethree = \{\varol{robot}{load}\}$. 
The state $\inpstatethree$ is also the final postcondition of $move_1$.
The skills are implemented by the robot's low-level controllers.
The skill abstraction can be obtained automatically~\cite{konidaris2018skills}. 
In this paper, we assume that the user provides the skill abstraction. 
We assume that 
skill execution does not terminate until it reaches the final postcondition, unless an 
assumption violation occurs, in which case we terminate the skill prematurely and start the recovery process.

\subsection{Synthesis and Specification Violation}
\label{sec:synth}
In this paper, we consider the \gls{gr1} fragment of \gls{ltl} to leverage its synthesis algorithm~\cite{bloem2012synthesis}. 
The specifications are of the form $\spec = \envspec \to \sysspec$ 
where $\envspec = \envinit \land \envsafety \land \envlive$ is the environment assumption and 
$\sysspec = \sysinit \land \syssafety \land \syslive$ is the system guarantee: 
\begin{flushitemize}
    \item $\envinit$ and $\sysinit$ are Boolean formulas over $\inp$ and $\out$, respectively, specifying the initial conditions. 
    \item $\envsafety$ and $\syssafety$ are the environment safety assumptions and the system safety guarantees, respectively. 
    They are of the form $\bigwedge_{i}\square\psi_i$ where $\psi_i$ is a Boolean formula over $v$ and $\bigcirc u$. 
    For $\envsafety$, $v \in \inp\cup\out$ and $u\in\inp$. 
    For $\syssafety$, $v, u \in \inp\cup\out$. 
    We divide $\envsafety$ into the skill assumptions $\envsafetynothard$ and the the hard assumptions $\envsafetyhard$ ($\envsafety = \envsafetynothard \land \envsafetyhard$). 
    The skill assumptions $\envsafetynothard$ describe the postconditions of the skills and can be modified by the repair process (see Section~\ref{sec:repair}). 
    The assumptions in $\envsafetyhard$ are the hard assumptions that cannot be modified by repair.
    \item $\envlive$ and $\syslive$ are the environment liveness assumptions and the system liveness goals, respectively. 
    They are of the form $\bigwedge_i\square\lozenge J_i$ where $J_i$ is a Boolean formula over $\inp\cup\out$. 
\end{flushitemize}

The synthesis algorithm in~\cite{bloem2012synthesis} formulates a two-player game between the environment and the system. 
The initial conditions of the specification define the initial positions of the players, 
the safety formulas define the transition relations of the players, 
and the liveness formulas induce the winning condition for the system. 
The synthesis algorithm solves a $\mu$-calculus formula with three nesting fixpoints
on the game structure to compute a 
winning strategy for the system. 

\noindent \textbf{\textit{Definition 1.}}
    A \textit{strategy} is a deterministic finite state machine $\strategy = (\Sigma, \Sigma_0, \inp, \out, \delta, \gamma_\inp, \gamma_\out)$ where 
    \begin{flushitemize}
        \item $\Sigma$ is a set of states and $\Sigma_0 \subseteq \Sigma$ is the set of initial states. 
        \item $\delta: \Sigma \times 2^\inp \rightharpoonup \Sigma$ is the transition function that maps a state and an input state to the next state. 
        \item $\gamma_\inp: \Sigma \to 2^\inp$ maps a state to its input state.
        \item $\gamma_\out: \Sigma \to 2^\out$ maps a state to its output state. 
    \end{flushitemize}
%
%
For subset $V \subseteq \vars$, let $\bigcirc V \coloneqq \{\bigcirc \pi \mid \pi\in V\}$. 
We define the violation of safety assumptions during strategy execution:

\noindent \textbf{\textit{Definition 2.}}
    Given the safety formula $\square\psi$ 
    where $\psi$ is a Boolean formula over $\inp \cup \out \cup \bigcirc \inp$,
    and 
    a triplet $(\inpstate,\outstate,\inpstateprime)$ where $\inpstate,\inpstateprime\subseteq\inp$ and $\outstate\subseteq \out$,
    the triplet $(\inpstate,\outstate,\inpstateprime)$ \textit{violates} $\square\psi$ iff 
    $\inpstate\cup\outstate\cup\bigcirc\inpstateprime\not\models\psi$. 





\section{Problem Statement}
This paper addresses the problem of automatically recovering from assumption violations during the task execution. 

\noindent \textbf{\textit{Problem \problemlabel{problem:1}.}} Given 
\begin{enumerate*}[label=(\roman*),ref=\roman*]
    \item a discrete abstraction of the environment $\inp$,
    \item a set of robot skills $\out$, and 
    \item a GR(1) task $\spectask$;   
\end{enumerate*} 
at runtime,
detect violations of the environment safety assumptions  $\square\psi_i$ in $\envsafety = \bigwedge_i\square\psi_i$ and repair the task, if possible.

\begin{algorithm}[ht]
\caption{\textbf{Strategy Execution and Recovery}}\label{alg:execution}
\KwIn{
Spec. $\spectask$,
Inverse grounding function $\inversegrounding$
        }
\KwOut{$\false$ if unrecoverable, otherwise the execution loop runs indefinitely}
$\strategy \coloneqq \tt{synthesize}(\spectask)$\;\label{line:execution_synth}
$\monitor \coloneqq \tt{monitor\_compiler(\spectask)}$\;\label{line:execution_monitor}
\tcp{Execution loop}
\While{$\true$}{\label{line:execution_loop}
    $\inpstate \coloneqq \tt{get\_inputs}(\inversegrounding)$\;\label{line:get_inputs}
    $\outstate \coloneqq \tt{get\_outputs}(\strategy, \inpstate)$\;\label{line:get_outputs}
    $\tt{execute}(\outstate)$\;\label{line:execute_outputs}
    \While{$\neg \tt{finished}(\outstate) \land (\inpstate = \tt{get\_inputs}(\inversegrounding))$}{\label{line:wait_start}
    \textbf{continue}\;
    }\label{line:wait_end}
    $\inpstateprime \coloneqq \tt{get\_inputs}(\inversegrounding)$\;\label{line:get_inputs_prime}
    $\cI \coloneqq \tt{check\_assump}(\monitor, (\inpstate, \outstate, \inpstateprime))$\;\label{line:check_assumptions}
    \If{$\cI \neq \emptyset$}{
        $\spectask \coloneqq \tt{relax}(\spectask, \cI, (\inpstate, \outstate, \inpstateprime))$\;\label{line:relax}
        \If{$\tt{unrealizable}(\spectask)$}{
        $\outnew \coloneqq \tt{repair}(\spectask)$\;\label{line:repair}
        \If{$\outnew \coloneqq \emptyset$}{\label{line:empty_outnew}
        \Return $\false$;
        }\label{line:empty_outnew_end}
        $\spectask \coloneqq \tt{add\_skill}(\spectask, \outnew)$\;\label{line:add_skills}
        $\strategy \coloneqq \tt{synthesize}(\spectask)$\;\label{line:resynth}
        $\monitor \coloneqq \tt{monitor\_compiler(\spectask)}$\;\label{line:re_monitor}
        }
    }
    
}\label{line:execution_end}
\end{algorithm}
\section{Approach}\label{sec:approach}
To solve Problem~\ref{problem:1},
while the robot is executing strategy $\strategy$ synthesized from $\spectask$,
we check whether 
the strategy execution 
violates any environment safety assumption at every timestep. 
If so, we relax the violated assumptions so that 
the execution no longer violates them, 
and we allow the robot to resume the execution from the current state. 
If the modified specification is no longer realizable, we find a set of new skills to make it realizable.

Algorithm~\ref{alg:execution} describes our approach. 
Given a specification $\spectask$, we use Slugs~\cite{ehlers2016slugs} to synthesize a strategy $\strategy$
and we generate a monitor $\monitor$ (Section~\ref{sec:detection}) 
that checks whether the safety assumptions are violated (lines~\ref{line:execution_synth}-\ref{line:execution_monitor}). 
We then execute the strategy in an execution loop (lines~\ref{line:execution_loop}-\ref{line:execution_end}).  


For each iteration of the execution loop, 
an inverse grounding function $\inversegrounding$ first gives the current input state $\inpstate$ (line~\ref{line:get_inputs}). 
Given $\inpstate$, the strategy $\strategy$ produces an output state $\outstate$ (line~\ref{line:get_outputs}). 
In lines~\ref{line:execute_outputs}-\ref{line:wait_end},
we execute the robot skills in $\outstate$ 
and wait for either the skills to complete the execution
(i.e. the robot completed a transition in $\strategy$)
or the input state to change
(i.e. the environment changed and we may need to change which skill the robot is activating). 
Next, $\inversegrounding$ gives the current input state $\inpstateprime$ (line~\ref{line:get_inputs_prime}). 
Then in line~\ref{line:check_assumptions}, the monitor $\monitor$ checks whether the triplet $(\inpstate, \outstate, \inpstateprime)$ violates any safety assumption and returns a set of assumptions $\cI$ that are violated by the triplet (Section~\ref{sec:detection}). 
If $\cI$ is empty, the program jumps to line~\ref{line:execution_loop} and starts a new iteration of the execution loop. 
If $\cI$ is nonempty, in line~\ref{line:relax} we relax the violated assumptions in $\cI$ and re-synthesize a strategy (Section~\ref{sec:relaxation}). 
If synthesis fails, in line~\ref{line:repair} we leverage synthesis-based repair~\cite{pacheck2020finding, pacheck2022automatic, pacheck2022physically}
to find a set of new skills $\outnew$ that allows the robot to complete the tasks (Section~\ref{sec:repair}). 
If repair cannot find any skills, i.e. $\outnew = \emptyset$, the execution terminates and we provide feedback to the user (lines~\ref{line:empty_outnew}-\ref{line:empty_outnew_end}). 
Otherwise, 
in lines~\ref{line:add_skills}-\ref{line:re_monitor}
we add $\outnew$ to the specification, generate a new strategy and the corresponding monitor, and go to line~\ref{line:execution_loop} to start a new iteration.
\subsection{Violation Detection}
\label{sec:detection}

The violation detection procedure 
takes in a specification $\spectask$
and
a triplet $(\inpstate,\outstate,\inpstateprime)$ where $\inpstate,\inpstateprime\subseteq\inp$ and $\outstate\subseteq\out$.
We use a monitor, compiled from $\spectask$, to detect violations of environment safety assumptions. 

\begin{figure}[t]
    \centering
    \includegraphics
    [width=0.33\textwidth]{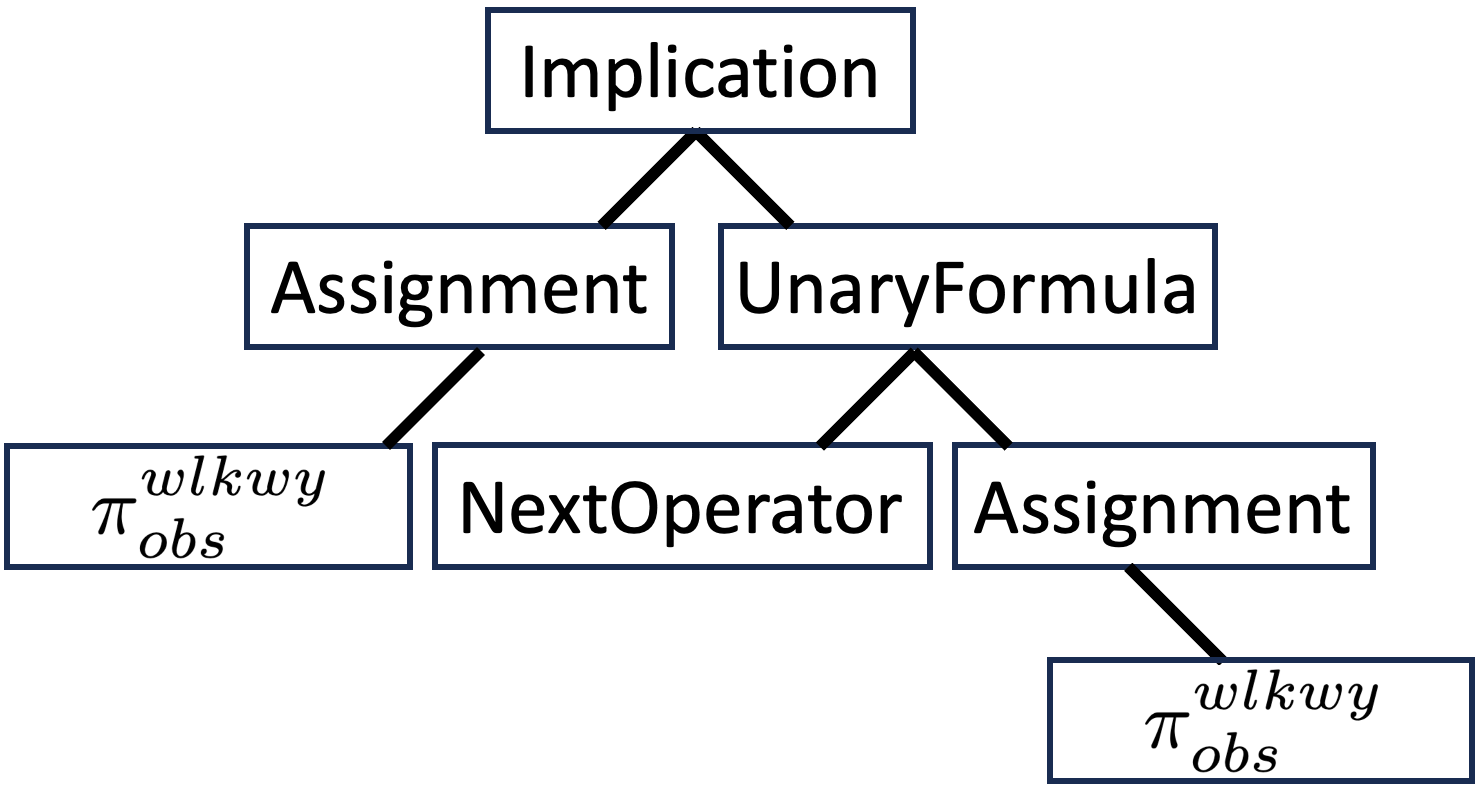}
    \caption{Abstract Syntax Tree for $\varol{obs}{wlkwy} \to \bigcirc \varol{obs}{wlkwy}$. 
    }
    \label{fig:ast}
\end{figure}

We use a compiler to translate the environment safety assumptions $\envsafety = \bigwedge_{i} \square\psi_i$ in $\spectask$ into a monitor $\monitor$. 
The 
compiler first uses a tokenizer and a parser to translate $\envsafety$ into a set of Abstract Syntax Trees (ASTs)~\cite{alfred2007compilers}. 
Each AST represents a Boolean formula $\psi_i$ 
in a tree structure according to the syntax of the specifications~\cite{ehlers2016slugs}. 
Then 
the compiler translates the ASTs to a set of Boolean constraints for the SAT solver Z3~\cite{de2008z3},
which is then used to check whether any safety assumptions are violated.
In Example~\ref{example:1}, the compiler translates 
$\varol{obs}{wlkwy} \to \bigcirc \varol{obs}{wlkwy}$
to the AST in Fig.~\ref{fig:ast} and creates the Boolean expression $\tt{Implies}$$(\varol{obs}{wlkwy}, \bigcirc\varol{obs}{wlkwy})$ in the Z3 format. 

The monitor $\monitor$ uses the triplet $(\inpstate, \outstate, \inpstateprime)$ to evaluate the environment safety assumption $\envsafety$. 
For each safety assumption $\square\psi_i$, 
the monitor evaluates the truth value of the Boolean formula $\psi_i$ 
given the literal values indicated by the triplet $(\inpstate,\outstate,\inpstateprime)$.
In practice, the monitor creates the following Boolean formula for each assumption $\square\psi_i$: 
\begin{equation}
    \label{eq:monitor}
    \begin{split}
        \psicheck \coloneqq \psi_i \land 
        \bigwedge_{x\in \inpstate} x \land \bigwedge_{x\in \inp \setminus \inpstate} \neg x \land 
        \bigwedge_{y\in \outstate} y \land \bigwedge_{y\in \out \setminus \outstate} \neg y \land \\
        \bigwedge_{x\in \inpstateprime} \bigcirc x \land \bigwedge_{x\in \inp \setminus \inpstateprime} \neg \bigcirc x
    \end{split}
\end{equation}
The monitor uses the SMT solver
Z3~\cite{de2008z3} to compute the satisfiability of Formula~\ref{eq:monitor}, which indicates whether the triplet violates the assumption $\square\psi_i$ or not. 
The monitor returns a set of violated assumptions 
$\cI \coloneqq \{\square\psi_i \mid (\inpstate,\outstate,\inpstateprime) \text{ violates } \square\psi_i\}$. 

In Example~\ref{example:1}, when the obstacle moves from \textit{Walkway} to \textit{Aisle}, 
$\inpstate = \{\varol{robot}{assm}, \varol{obs}{wlkwy}\}$, 
$\outstate = \{move_1\}$, 
$\inpstateprime = \{\varol{robot}{assm}, \varol{obs}{aisle}\}$.
For $\psi_i = \varol{obs}{wlkwy} \to \bigcirc \varol{obs}{wlkwy}$, $\psicheck$ in Formula~\ref{eq:monitor} is unsatisfiable, 
therefore $\cI = \{\square(\varol{obs}{wlkwy} \to \bigcirc \varol{obs}{wlkwy})\}$.





\subsection{Assumption Relaxation}
\label{sec:relaxation}
The assumption relaxation procedure takes in
    a specification $\spectask$, 
    a set of violated environment safety assumptions $\cI$, 
    and a triplet $(\inpstate, \outstate, \inpstateprime)$ 
    that violates the assumptions in $\cI$.
The procedure relaxes the violated assumptions in $\cI$ to admit the violated transition indicated by the triplet. 
The procedure follows \gls{ec} in~\cite{wong2014correct, wong2018resilient}, which relaxes the assumptions solely using input states.
In contrast to \gls{ec}, our procedure utilizes 
the complete observation triplet $(\inpstate, \outstate, \inpstateprime)$ to rewrite violated assumptions.
Consequently, our procedure can handle the discovery of previously unknown nondeterminism in robot skills by
incorporating the nondeterminism into the skill abstraction.
In addition, our procedure allows the robot to resume the execution in the current input state and continue to attempt the latest liveness goal.

Recall that we split the environment safety assumptions $\envsafety = \bigwedge_i \square\psi_i$ into the skill assumptions $\envsafetynothard$,
which describe the postconditions of the skills,
and the hard assumptions $\envsafetyhard$, which cannot be modified by the repair procedure in Section~\ref{sec:repair}. 
For each violated assumption $\square\psi_i$ in $\envsafetyhard$, 
the procedure replaces $\square\psi_i$ with $\square(\psi_i\lor\envrelaxhard)$ 
where $\envrelaxhard$ is the following formula:
\begin{equation}
    \label{eqn:relax_hard}
    \begin{split}
        \envrelaxhard \coloneqq
        \bigwedge_{x\in \inpstate} x \land \bigwedge_{x\in \inp \setminus \inpstate} \neg x \land 
        \bigwedge_{y\in \outstate} y \land \bigwedge_{y\in \cY \setminus \outstate} \neg y \land \\
        \bigwedge_{x\in \inpstateprime} \bigcirc x \land \bigwedge_{x\in \inp \setminus \inpstateprime} \neg \bigcirc x  
    \end{split}
\end{equation}     
For each violated assumption $\square\psi_i$ in $\envsafetynothard$, 
the procedure replaces the assumption $\square\psi_i$ with 
$\square(\psi_i\lor\envrelaxnothard)$
where $\envrelaxnothard$ is shown in Formula~\ref{eqn:relax_not_hard}.
Intuitively, both formulas add the observed behavior, using disjunction, to the assumptions, thereby allowing for more environment behaviors. 
The difference between Formula~\ref{eqn:relax_not_hard} and~\ref{eqn:relax_hard} is that the former only adds the controllable inputs, and the latter adds all inputs. 
The reason for this distinction is that skill abstractions are defined solely over controllable inputs,
so we can only modify them with controllable inputs,
whereas the hard assumptions $\envsafetyhard$ are defined over all inputs.
\begin{equation}
\label{eqn:relax_not_hard}
\begin{split}
    \envrelaxnothard \coloneqq
    \bigwedge_{x\in \inpstate \cap \inpsyms} x \land \bigwedge_{x\in \inpsyms \setminus \inpstate} \neg x \land
    \bigwedge_{y\in \outstate} y \land \bigwedge_{y\in \cY \setminus \outstate} \neg y \land \\
    \bigwedge_{x\in \inpstateprime \cap \inpsyms} \bigcirc x \land \bigwedge_{x\in \inpsyms \setminus \inpstateprime} \neg \bigcirc x
\end{split}
\end{equation}
\begin{figure}
    \centering
    \includegraphics
    [width=0.46\textwidth]{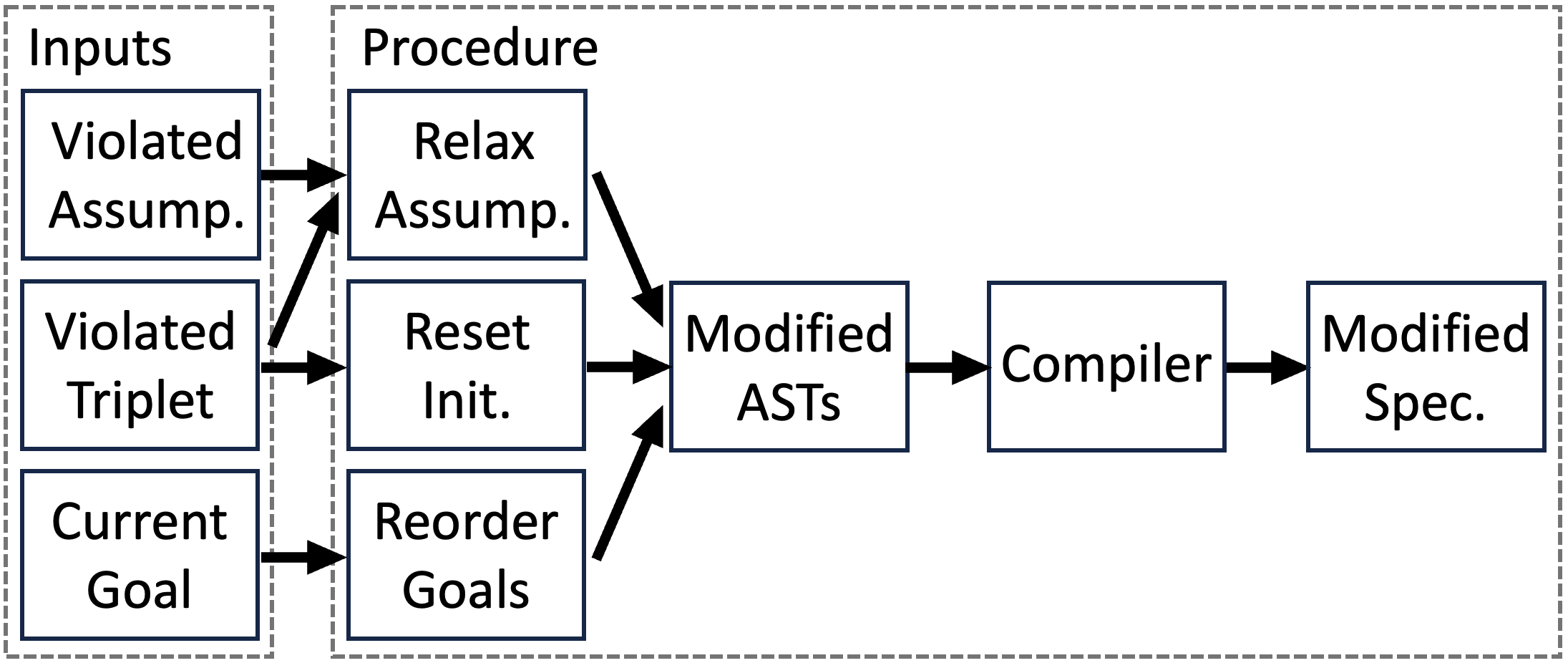}
    \caption{Workflow of the assumption relaxation procedure.}
    \label{fig:relaxation}
\end{figure}

Fig.~\ref{fig:relaxation} depicts the workflow of the relaxation procedure in practice. 
We relax the violated assumptions by modifying the ASTs in the monitor compiler (Section~\ref{sec:detection}) according to Formulas~\ref{eqn:relax_hard} and \ref{eqn:relax_not_hard}. 
The compiler then generates the new specification by recursively visiting every node in the modified ASTs and generating the corresponding \gls{ltl} formula. 

Before generating the new specification, we set the initial input condition $\envinit$ to be the current input state $\inpstateprime$, as shown in Formula~\ref{eqn:init_state}. 
Doing so allows the robot to resume the task execution in the current input state. 
\begin{equation}\label{eqn:init_state}
    \envinit \coloneqq \bigwedge_{x\in \inpstateprime} x \land \bigwedge_{x\in \inp\setminus\inpstateprime} \neg x
\end{equation}

Moreover, we reorder the liveness goals to 
allow the robot to attempt 
the latest goal first. 
Recall the liveness goal formula $\syslive = \bigwedge_{i=1}^m \square\lozenge J_i$. 
Suppose the strategy attempts to satisfy the $j^{\textrm{th}}$ liveness goal $J_j$ before the violation happens.
Since the synthesized strategy loops through the liveness goals in a cyclic manner~\cite{bloem2012synthesis}, 
we reorder $\syslive$ by moving $J_j \dots J_m$ to the front of $\syslive$.


\subsection{Repair}
\label{sec:repair}
The repair procedure takes in 
    an unrealizable specification $\spectask$ and
    the abstraction $(\inp, \out)$. 
We find a set of new skills $\outnew$ and modify the specification $\spectask$ based on the new abstraction $(\inp, \out\cup\outnew)$ such that the new specification includes the new skills $\outnew$ and is realizable. 

We leverage the synthesis-based repair approach proposed in~\cite{pacheck2020finding, pacheck2022automatic, pacheck2022physically} to find a set of new skills. 
The algorithm suggests new skills by alternating two procedures: 
\begin{enumerate}
    \item \label{ls:modify_pre}\textit{Modify-Preconditions.} 
    Randomly select an existing skill transition such that the robot can satisfy the task from its postcondition.
    Modify the precondition of the transition to a random new state from which the robot cannot satisfy the task before the modification.
    \item \label{ls:modify_post}\textit{Modify-Postconditions.} 
    Randomly select an existing skill transition 
    such that the robot cannot satisfy the task from either the pre- or postcondition of the transition. 
    Modify the postcondition to a random new state from which the robot can satisfy the task.
\end{enumerate}



In this work we change the Modify-Preconditions algorithm from~\cite{pacheck2020finding, pacheck2022automatic, pacheck2022physically} 
to improve the efficiency of repair. 
Repair can be inefficient
since both modifying the pre- and postconditions involves randomly selecting a transition and a new state. 
In Modify-Precondition, 
suppose the original precondition is an intermediate state of a skill. Repair ensures that the new precondition is reachable from the initial precondition of the skill by adding a transition from the precondition of the original precondition (pre-precondition) to the new precondition. 
However, even if the robot can satisfy the task from the new precondition after the modification, the robot still cannot satisfy the task from the pre-precondition because it has a nondeterministic transition to the old precondition. 
We remove the transition from the pre-precondition to the old precondition, allowing the robot to satisfy the task from the pre-precondition. 
In Example~\ref{example:1}, the original Modify-Precondition algorithm returns a symbolic skill $move'$ which goes from $\varol{robot}{assm}$ nondeterministically to either $\varol{robot}{wlkwy}$ or $\varol{robot}{aisle}$, and from $\varol{robot}{wlkwy}$ to $\varol{robot}{load}$. 
Because of the nondeterminism, the robot cannot guarantee it will reach \textit{Loading} and requires further refinement of the skills.
Our approach returns the skill $move_2$,
which goes from
$\varol{robot}{assm}$ 
to $\varol{robot}{wlkwy}$ 
to $\varol{robot}{load}$,
in one iteration of the repair. 


Given a set of suggested skills $\outnew$ from the repair algorithm, in order to decrease the size of the abstraction, 
we attempt to remove unnecessary skills in $\outnew$.
For each skill $y\in\outnew$, we remove $y$ from $\outnew$
and check whether the specification with the updated $\outnew$ is realizable.
If the result is realizable, then we remove the skill $y$
permanently. 
Otherwise, we add the skill $y$ back to the set of new skills.

We physically implement the new symbolic skills using a motion planner~\cite{kavraki1996probabilistic}.
If we cannot physically implement a symbolic skill, 
we incorporate the skill as a repair constraint,
as done in~\cite{pacheck2022physically}, and continue the repair process.

\begin{figure}
    \centering
    \includegraphics[width=0.449\textwidth]{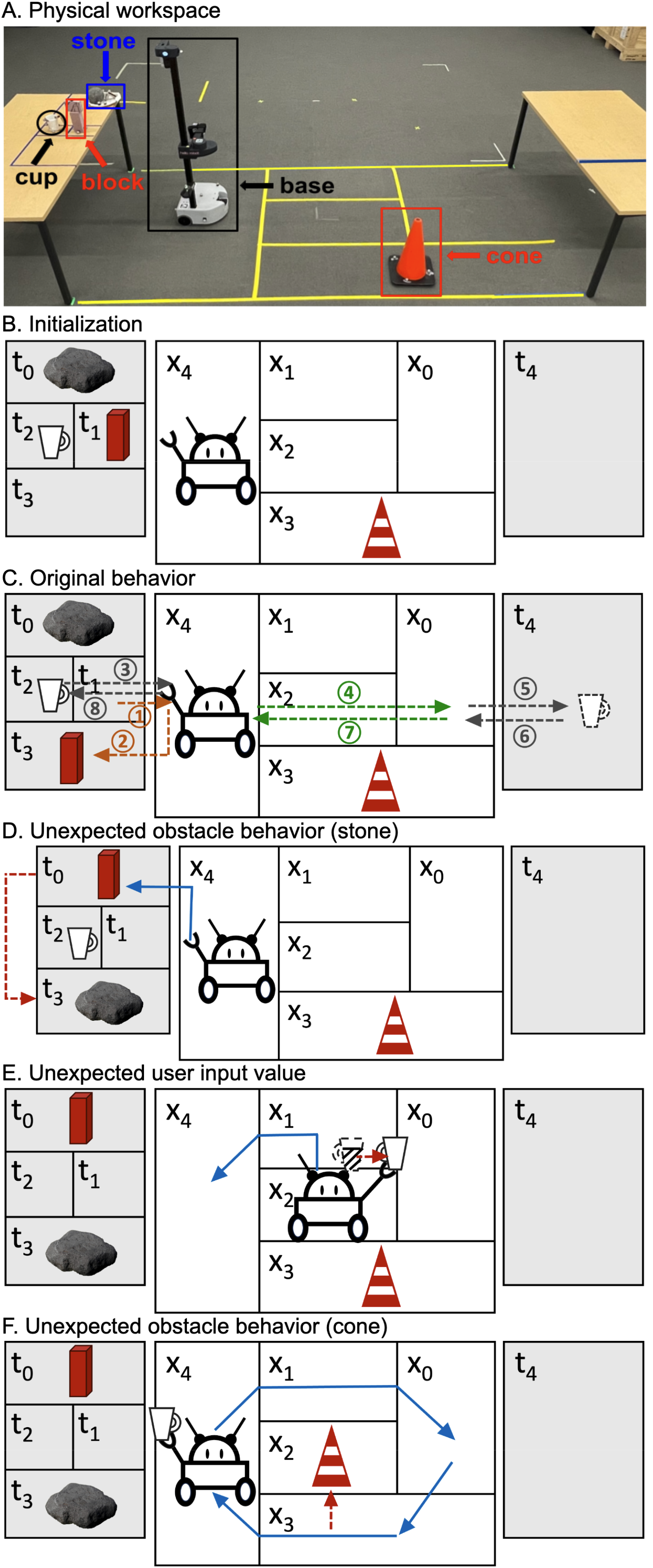}
    \caption{Stretch moving $\tt{cup}$ between tables. 
    The task is $\square\lozenge (\varol{empty}{}\to\varol{cup}{t_2}) \land 
    \square\lozenge (\neg\varol{empty}{}\to\varol{cup}{t_4})$.
    (A) Physical workspace.
    (B) Initial positions of the objects.
    (C) Original behavior of the synthesized strategy. 
    Arrows represent the motion of the objects, 
    orange: $\tt{block}$, 
    grey: $\tt{cup}$, 
    green: $\tt{base}$.
    The numbers represent the order of their movements.
    (D) Recovery from Violation 1 where $\tt{stone}$ moves from $t_0$ to $t_3$.
    (E) Recovery from Violation 2 where $\tt{cup}$ becomes empty in $x_2$.
    (F) Recovery from Violation 3 where $\tt{cone}$ moves from $x_3$ to $x_2$.
    In (D)-(F), red dashed arrows represent the violations and blue solid arrows represent the new skills.}
    \label{fig:demo}
\end{figure}


\section{Demonstration}
We demonstrate our approach by employing a Hello Robot Stretch~\cite{kemp2022design} in a factory-like setting.  
We present the abstraction, the tasks, and three assumption violations and their recoveries within one run of the task execution.  
We perform the computations (synthesis and repair) on a Linux desktop 
with a 2.3GHz 8-core AMD Ryzen\textsuperscript{TM}-7 CPU. 
We run our framework on the desktop and send control inputs to Stretch through ROS. 
We open-source the code for our framework and demonstration\footnote{\url{https://github.com/MartinMeng008/env_repair}}.
A video of the demonstration is available at: 
\url{https://youtu.be/OTUEyqQfQQs}.   

\subsection{Setting}\label{sec:demo_setting}
\textbf{\textit{Example~\exampletwolabel{example:2}.}}
The robot is asked to repeatably bring a cup to a loading table if the cup is empty and bring the cup to an assembly table if the cup is full, while avoiding collisions with the 
block, the stone, and the cone
(see Fig.~\ref{fig:demo}A).

\textbf{\textit{Abstraction.}}
We partition the continuous workspace into a set of rectangular regions 
    $\cR \coloneqq \{
    x_0, \dots, x_4, 
    t_0, \dots, t_4\}$,
    where $x_i$ represents a region on the ground, 
    and $t_i$ represents a region on a table,
    as illustrated in Fig.~\ref{fig:demo}B.
    The set of objects $\cO$ includes the robot $\tt{base}$, the controllable $\tt{cup}$, the controllable obstacle $\tt{block}$, and the uncontrollable obstacles $\tt{cone}$ and $\tt{stone}$  (see Fig.~\ref{fig:demo}A). 
    We represent the state of the cup (empty or full) using $\varol{empty}{}$. 
    For the objects that can be moved by the robot ($\tt{cup,block}$), 
    we add propositions indicating whether they are in the robot end effector or not ($\varol{cup}{ee},\varol{block}{ee}$).
    We define the set of input propositions as $\inp \coloneqq \{\varol{o}{r} \mid o \in \cO, r \in \cR \} \cup \{\varol{empty}{},\varol{cup}{ee},\varol{block}{ee}\}$. 
    We ground each input $\varol{o}{r}$ 
    by $\grounding(\varol{o}{r}) \coloneqq \{x \mid \tt{pose}(x, o) \in \tt{region}(r)\}$.
    For $o \in \{\tt{cup}, \tt{block}\}$, 
    $\grounding(\varol{o}{ee}) \coloneqq \{x \mid \norm{\tt{pose}(x, o) - \tt{default\_pose}(x, ee)} < \epsilon\}$ for some small $\epsilon$.
    We use an OptiTrack motion capture system to determine the positions of the objects and the robot. 
    We control the value of input $\varol{empty}{}$, indicating whether the cup is empty or not, through a keyboard interface.

\textbf{\textit{Skills.}}
The robot has two mobile skills:
    $\tt{move_1}$ goes from $\varol{base}{x_0}$ to $\varol{base}{x_2}$ to $\varol{base}{x_4}$, and
    $\tt{move_2}$ goes from $\varol{base}{x_4}$ to $\varol{base}{x_2}$ to $\varol{base}{x_0}$,
    while other controllable inputs remain unchanged. 
    The robot has six manipulation skills:
    $\tt{pick_1}$ goes from $\varol{cup}{t_2}$ to $\varol{cup}{ee}$, 
    $\tt{pick_2}$ goes from $\varol{cup}{t_4}$ to $\varol{cup}{ee}$, 
    $\tt{pick_3}$ goes from $\varol{block}{t_1}$ to $\varol{block}{ee}$, 
    $\tt{place_1}$ goes from $\varol{cup}{ee}$ to $\varol{cup}{t_2}$, 
    $\tt{place_2}$ goes from $\varol{cup}{ee}$ to $\varol{cup}{t_4}$, 
    and $\tt{place_3}$ goes from $\varol{block}{ee}$ to $\varol{block}{t_3}$, 
    while other controllable inputs remain unchanged. 

\textbf{\textit{Tasks.}}
The task is 
    $\syslive =
    \square\lozenge (\varol{empty}{}\to\varol{cup}{t_2}) \land 
    \square\lozenge (\neg\varol{empty}{}\to\varol{cup}{t_4})$; 
    when the cup is empty,
    the robot should place it on the loading table $t_2$, 
    and when the cup is full, the robot should place it on the assembly table $t_4$. 

\textbf{\textit{Initialization.}}
As shown in Fig.~\ref{fig:demo}B, 
    the robot starts in $x_4$ ($\varol{base}{x_4}$), 
    the uncontrollable $\tt{cone}$ is in $x_3$ ($\varol{cone}{x_3}$),
    the uncontrollable $\tt{stone}$ is in $t_0$ ($\varol{stone}{t_0}$), 
    the controllable 
    $\tt{block}$ is in $t_1$ ($\varol{block}{t_1}$), 
    and the controllable $\tt{cup}$ is in $t_2$ ($\varol{cup}{t_2}$).  
    The cup is initially empty ($\varol{empty}{}$). 

\textbf{\textit{Constraints.}}
The robot is required to satisfy the following safety constraints during the task execution:
\begin{enumerate}[label=\arabic*),ref=\arabic*]
    \item \label{enu:const_1} A region can contain only one object at a time: 
    $\square \neg(\varol{o_1}{r} \land \varol{o_2}{r}) \land 
    \square \neg(\bigcirc \varol{o_1}{r} \land \bigcirc\varol{o_2}{r})$ 
    for any objects $o_1, o_2 \in \cO$ and region $r \in \cR \cup \{ee\}$.
    \item \label{enu:const_2}Since $\tt{block}$ is tall, the robot cannot pick up $\tt{cup}$ from $t_2$ if $\tt{block}$ is in $t_1$: 
    $\square (\varol{block}{t_1} \land \varol{cup}{t_2} \to \neg \bigcirc \varol{cup}{ee})$. 
    \item \label{enu:const_3}We do not allow the robot to enter the assembly area $x_0$ with an empty cup, nor enter the loading area $x_4$ with a full cup: 
    $\square (\varol{cup}{ee} \land \varol{empty}{} \land \neg \varol{base}{x_0} \to \neg \bigcirc \varol{base}{x_0}) \land \square (\varol{cup}{ee} \land \neg \varol{empty}{} \land \neg \varol{base}{x_4} \to \neg \bigcirc \varol{base}{x_4})$. 
\end{enumerate}

\textbf{\textit{Assumptions.}}
We make the following environment safety assumptions to ensure the specification is realizable:
\begin{enumerate}[label=\arabic*),ref=\arabic*]
    \item \label{assume:1}We assume that the uncontrollable obstacles are static: 
    $\square (\varol{o}{r} \to \bigcirc \varol{o}{r})$ 
    for $o\in\{\tt{cone},\tt{stone}\}$ and $r \in \cR $. 
    \item \label{assume:2} We assume that if the status of the cup changed from empty to full then it was in $t_2$, and if it changed from full to empty it was in $t_4$: 
    $\square (\varol{empty}{} \land \neg \bigcirc \varol{empty}{} \to \varol{cup}{t_2}) \land 
    \square (\neg\varol{empty}{} \land \bigcirc \varol{empty}{} \to \varol{cup}{t_4})$.
\end{enumerate}

\textbf{\textit{Behavior.}}
We synthesize a strategy and execute it with the Stretch robot,
as illustrated in Fig.~\ref{fig:demo}C.
    Initially, while $\tt{cup}$ is empty, the robot does not move. 
    When $\tt{cup}$ is full, the robot first executes 
    $\tt{pick_3}$ and $\tt{place_3}$ to pick up $\tt{block}$ from $t_1$ and place it in $t_3$. 
    Then the robot executes 
    $\tt{pick_1}$, $\tt{move_2}$, and $\tt{place_2}$ sequentially to pick up $\tt{cup}$ from $t_2$, move to $x_2$ and then $x_0$, and place $\tt{cup}$ in $t_4$. 
    While $\tt{cup}$ remains full the robot does not move. 
    Once $\tt{cup}$ becomes empty, the robot executes 
    $\tt{pick_2}$, $\tt{move_1}$, and $\tt{place_1}$ sequentially to 
    pick up $\tt{cup}$ from $t_4$, move to $x_2$ and then $x_4$, and finally place $\tt{cup}$ back in $t_2$.
    This behavior continues indefinitely. 


\subsection{Unexpected Obstacle Behavior - Stone}
\textbf{\textit{Violation 1.}}
In Fig.~\ref{fig:demo}D, 
after $\tt{cup}$ is full and the robot starts to execute $\tt{pick_3}$, the uncontrollable $\tt{stone}$ is moved from $t_0$, violating Assumption~\ref{assume:1} ($\tt{stone}$ is static). 
    The previous input state is
    $\inpstate = \{$
    $\varol{cup}{t_2},$
    $\varol{base}{x_4},$ 
    $\varol{cone}{x_3},$
    $\varol{block}{t_1},$
    $\varol{stone}{t_0}\}$, 
    the previous output state is 
    $\outstate = \{\tt{pick_3}\}$, 
    and the current input state is 
    $\inpstateprime = \{\varol{cup}{t_2},$
    $\varol{base}{x_4},$
    $\varol{cone}{x_3},$
    $\varol{block}{t_1},$
    $\varol{stone}{t_3}\}$.  

The monitor detects that the triplet $(\inpstate, \outstate, \inpstateprime)$ 
    violates the assumption
    $\envsafetyvio = \square (\varol{stone}{t_0} \to \bigcirc \varol{stone}{t_0})$ and terminates the execution of $\tt{pick_3}$. 
%
The relaxation procedure took 0.4 seconds to relax $\envsafetyvio$ with the triplet $(\inpstate, \outstate, \inpstateprime)$ according to Formula~\ref{eqn:relax_hard}. 
%
The specification becomes unrealizable because $\tt{stone}$ is in $t_3$, preventing the robot from executing $\tt{place_3}$ to place $\tt{block}$ in $t_3$. 
The repair took 16 seconds to suggest a skill $\tt{place_{4}}$ that goes from $\varol{block}{ee}$ to $\varol{block}{t_0}$ while other controllable inputs remain unchanged.

The framework took 1.1 seconds to synthesize a new strategy and the robot follows the strategy to execute $\tt{pick_3}$ and $\tt{place_{4}}$ sequentially to pick up $\tt{block}$ and place it in $t_0$. 
        The robot then executes $\tt{pick_1}$ to pick up $\tt{cup}$ from $t_2$ and resumes the behavior as described in Section~\ref{sec:demo_setting}. 

\subsection{Unexpected User Input Change}
\textbf{\textit{Violation 2.}}
In Fig.~\ref{fig:demo}E, 
After the robot picks up $\tt{cup}$, the robot executes $\tt{move_2}$ to move from $x_4$ to $x_0$ through $x_2$. 
    When the robot is in $x_2$, $\tt{cup}$ becomes empty, violating Assumption~\ref{assume:2} (the status of $\tt{cup}$ can only change when $\tt{cup}$ is in $t_2$ or $t_4$). 
    The previous input state is
    $\inpstate = \{$
    $\varol{cup}{ee},$
    $\varol{base}{x_2},$ 
    $\varol{cone}{x_3},$
    $\varol{block}{t_0},$
    $\varol{stone}{t_3}\}$, 
    the previous output state is 
    $\outstate = \{\tt{move_2}\}$, 
    and the current input state is 
    $\inpstateprime = \{
    \varol{cup}{ee},$
    $\varol{base}{x_2},$
    $\varol{cone}{x_3},$
    $\varol{block}{t_0},$
    $\varol{stone}{t_3},$
    $\varol{empty}{}\}$. 

The monitor detects that the triplet $(\inpstate, \outstate, \inpstateprime)$ 
    violates the assumption 
    $\envsafetyvio =$
    $\square (\neg\pi_{empty} \land \bigcirc \pi_{empty} \to \varol{cup}{t_4})$ 
    and terminates the execution of $\tt{move_2}$.
The relaxation procedure took 0.4 seconds to relax the violated assumption $\envsafetyvio$ with $(\inpstate, \outstate, \inpstateprime)$ according to Formula~\ref{eqn:relax_hard}.
The specification becomes unrealizable because the robot is in $x_2$ with no existing skill having an initial precondition in $x_2$, 
and the robot cannot continue executing $\tt{move_2}$ to reach $x_0$, because bringing the empty cup to $x_0$ violates Constraint~\ref{enu:const_3}. 
The repair took 10.7 seconds to suggest two skills: 
$\tt{move_3}$ goes from $\varol{base}{x_2}$ to $\varol{base}{x_1}$ and 
$\tt{move_4}$ goes from $\varol{base}{x_1}$ to $\varol{base}{x_4}$ while other controllable inputs remain unchanged. 
In this particular run, the suggestion is not minimal due to the inherent randomness in repair. 
In another run, the repair took 38 seconds to provide a minimal suggestion containing only a single skill that goes from $\varol{base}{x_2}$ to $\varol{base}{x_4}$.

The framework took 1.4 seconds to synthesize a new strategy
in which the robot executes $\tt{move_3}$, $\tt{move_4}$ and $\tt{place_1}$ sequentially to move from $x_2$ to $x_1$ and then $x_4$, and place $\tt{cup}$ back to $t_2$. 
The robot does not move until $\tt{cup}$ becomes full again and resumes the execution as described in Section~\ref{sec:demo_setting}. 

\subsection{Unexpected Obstacle Behavior - Cone}

\textbf{\textit{Violation 3.}}
In Fig.~\ref{fig:demo}F, 
after $\tt{cup}$ becomes full and the robot executes $\tt{pick_1}$, 
    the robot holds the cup and executes $\tt{move_2}$ to drive from $x_4$ to $x_0$ through $x_2$. 
    When the robot is still in $x_4$, $\tt{cone}$ is moved from $x_3$ to $x_2$, violating Assumption~\ref{assume:1} ($\tt{cone}$ should be static). 
    The previous input state is 
    $\inpstate = \{$
    $\varol{cup}{ee},$
    $\varol{base}{x_4},$ 
    $\varol{cone}{x_3},$
    $\varol{block}{t_0},$
    $\varol{stone}{t_3}\}$, 
    the previous output state is 
    $\outstate = \{\tt{move_2}\}$, 
    and the current input state is 
    $\inpstateprime = \{$
    $\varol{cup}{ee},$
    $\varol{base}{x_4},$ 
    $\varol{cone}{x_2},$
    $\varol{block}{t_0},$
    $\varol{stone}{t_3}\}$. 

The monitor detects that the triplet $(\inpstate, \outstate, \inpstateprime)$ 
    violates the assumption
    $\envsafetyvio = \square (\varol{cone}{x_3} \to \bigcirc \varol{cone}{x_3})$ and terminates the execution of $\tt{move_2}$.
The relaxation procedure took 0.5 seconds to relax the violated assumption $\envsafetyvio$ with the triplet according to Formula~\ref{eqn:relax_hard}. 
The specification becomes unrealizable because 
Constraint~\ref{enu:const_1} prohibits the robot from entering $x_2$ and both $\tt{move_1}$ and $\tt{move_2}$ pass through $x_2$. 
The repair took 1 minute and 35 seconds to suggest three skills: 
$\tt{move_5}$ goes from $\varol{base}{4}$ to $\varol{base}{1}$ to $\varol{base}{0}$, 
$\tt{move_6}$ goes from $\varol{base}{0}$ to $\varol{base}{3}$, 
and $\tt{move_7}$ goes from $\varol{base}{3}$ to $\varol{base}{4}$, 
while other controllable inputs do not change. 

The framework took 2.6 seconds to synthesize
a new strategy which behaves the same as the original strategy (Section~\ref{sec:demo_setting}) except that it replaces
$\tt{move_2}$ with $\tt{move_5}$ and $\tt{move_1}$ with $\tt{move_6}$ and $\tt{move_7}$. 
The robot executes $\tt{move_5}$ and $\tt{place_2}$ sequentially to reach $x_0$ through $x_1$ and place $\tt{cup}$ in $t_4$. 
When $\tt{cup}$ becomes empty, the robot executes $\tt{pick_2}$, $\tt{move_6}$, $\tt{move_7}$, and $\tt{place_1}$ in sequence to pick up $\tt{cup}$ from $t_4$, reach $x_4$ through $x_3$, and place $\tt{cup}$ in $t_2$.

\section{Conclusion}
We provide a framework for automatically recovering from assumption violations of temporal logic specifications.
To do so we detect the assumption violations during the robot execution, relax the violated assumptions to admit the observed environment behavior, and add new robot skills for the robot to complete the tasks. 
We demonstrate our approach 
with a physical robot in a factory-like scenario. 

There are assumption violations, however, from which our approach cannot recover. 
If assumptions about the postconditions of skills are violated due to faulty low-level robot controllers, then repair at the symbolic level cannot solve the problem. 
In Example~\ref{example:1}, our framework cannot recover from the violation if the low-level controller cannot reliably implement the new skill $move_2$ due to severe localization error and uncertainty in \textit{Walkway}.
Moreover, if an assumption violation results in the robot's current state violating its hard safety constraints, 
then no new skills can repair the task from that violating state.
In Example~\ref{example:1}, if the obstacle moves to \textit{Aisle} when the robot is already in \textit{Aisle}, 
our approach cannot repair because the current state violates the safety constraint which prohibits the robot and the obstacle from being in the same region. 
In future work, we plan to 
tackle the violations that our approach cannot repair,
develop local repair to improve the computational efficiency of our approach,
and 
extend our framework to multi-agent and human-robot interaction settings where assumptions about human behaviors may be violated.

\bibliography{references}
\bibliographystyle{ieeetr}

\end{document}